\documentclass[conference,compsoc]{IEEEtran}
\usepackage{graphicx}
\usepackage{booktabs}

\ifCLASSOPTIONcompsoc
  \usepackage[nocompress]{cite}
\else
  \usepackage{cite}
\fi

\ifCLASSINFOpdf
\else
\fi
\begin{document}
%
\title{IoT Network Traffic Analysis with Deep Learning}

\author{\IEEEauthorblockN{Mei Liu}
\IEEEauthorblockA{School of IT\\
Deakin University\\
Australia\\
Email: merryliuny@gmail.com}
\and
\IEEEauthorblockN{Leon Yang}
\IEEEauthorblockA{School of IT\\
Deakin University\\
Australia\\
Email: leon.yang@deakin.edu.au}
}


%


\maketitle

\begin{abstract}
As IoT networks become more complex and generate massive amounts of dynamic data, it is difficult to monitor and detect anomalies using traditional statistical methods and machine learning methods. Deep learning algorithms can process and learn from large amounts of data and can also be trained using unsupervised learning techniques, meaning they don't require labelled data to detect anomalies. This makes it possible to detect new and unknown anomalies that may not have been detected before. Also, deep learning algorithms can be automated and highly scalable; thereby, they can run continuously in the backend and make it achievable to monitor large IoT networks instantly. In this work, we conduct a literature review on the most recent works using deep learning techniques and implement a model using ensemble techniques on the KDD Cup 99 dataset. The experimental results showcase the impressive performance of our deep anomaly detection model, achieving an accuracy of over 98\%.
\end{abstract}


%
\IEEEpeerreviewmaketitle

\section{Introduction}
Anomalies, namely novelties or outliers, are individuals that are far from the nominal data population. In general, an anomaly can be thought of as an observation or pattern that does not conform to the expected behaviour or follows the same patterns as the rest of the data. Anomaly Detection (AD), is the process of identifying unusual patterns in data that do not conform to a well-defined notion of normal data \cite{chandola2009anomaly}. The process involves learning the normal behaviour of the data and then identifying instances that deviate significantly from the model through statistical methods or machine learning techniques. For example, in a time series dataset, an anomaly might be a sudden change in the trend or a spike in the data that is not consistent with the rest of the series. In a classification problem, an anomaly could be an observation that does not fit into any defined classes or is significantly different from the other observations. 

Statistically, the sparsely distributed areas indicate that the probability of data occurring in a certain area is relatively low, where the data falling in can be considered to be anomalies. In Figure 1, we illustrate the anomalies in two-dimensional data space, where the clusters in blue indicate normal data and red points represent anomalies far from normal. Given a dataset X = \{X1, X2, ..., Xn\}, the feature dimension of each sample is D, $x_i \in R ^ D$. Deep Anomaly Detection (DAD) aims to learn a mapping function, which maps the original space to a new representation space $\phi(\cdot)$: $X\mapsto Z$, where $Z \in R^K (K \ll D)$. If the probability density of a sample in the dataset is less than the threshold, a small enough value, the sample is considered an anomaly and the anomaly score of the sample $\tau(\cdot)$ can be computed in the new space. Such sparse anomalies can be applicable in many areas by analysing activity patterns to detect anomalous behaviours, manage industrial resources, or ensure production security. 

\begin{figure}[h]
    \centering
    \includegraphics[width=0.5\textwidth]{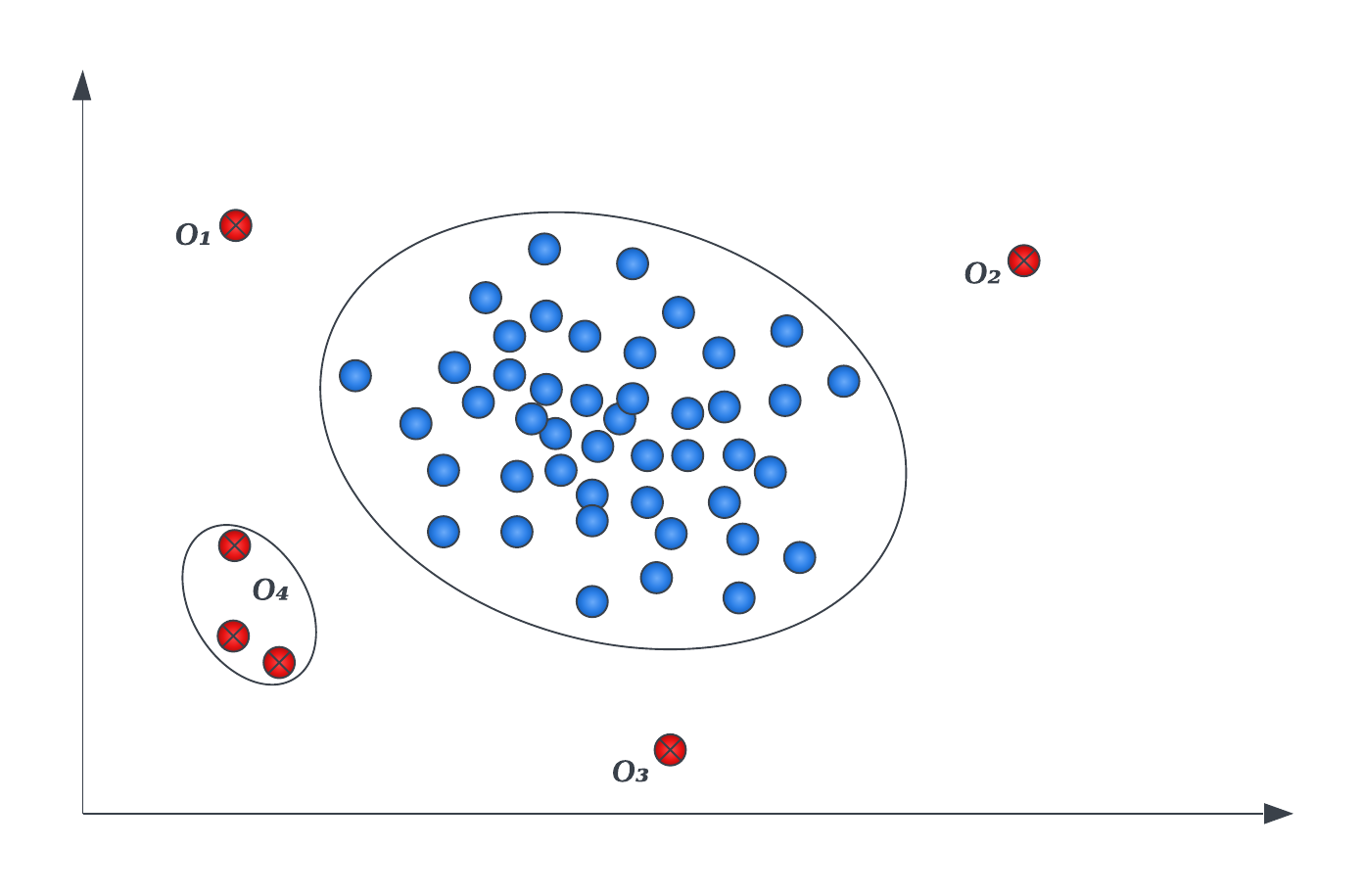}
    \caption{A simple example of anomalies in 2d data space as the red cross points while the blue solid points are the normal data.}
    \label{fig:Anomalies}
\end{figure}

Based on the availability of data labels, we usually divide Anomaly Detection tasks into three types: supervised, semi-supervised, and unsupervised. Supervised AD uses labels of nominal and anomalous data instances to train binary or multi-class classifiers; semi-supervised techniques use the existing normal single label to separate outliers without the anomalous instances involved in the training process; in training with unsupervised deep AD techniques, there are both normal and anomalous instances in the data, but the normal instances are often much larger than the anomalies. This method detects outliers based on the intrinsic properties of the data instances and is usually used for automatic labelling of unlabelled data samples. Even though denoting a data point with a label of normal or anomalous would be one of the ideal solutions for AD, it requires extreme effort to obtain the labelled training data. In addition, due to the nature of data, for example, the dynamic behaviours in anomalous pattern recognition would lead to further difficulties in labelling. In that case, a vital solution to AD is to train a normality model from normal data in an unsupervised manner to detect anomalies through deviations from the model.

In the context of the Internet of Things (IoT), where data is transmitted between IoT devices and systems over a network, anomalies are a special type of outlier point that usually carries meaningful pieces of information, such as sensor readings, device status and configuration data, and messages or commands sent between devices. The data format of IoT network traffic can vary depending on the specific devices and systems being used and the type of data being transmitted. Due to the complexity of IoT networks, identifying unexpected behaviours in the system is vital for preventing anomalies from escalating into larger issues. For example, AD can help identify potential security threats, such as unauthorised access to the system or unusual data usage patterns that could indicate a cyber-attack. Especially with the continuous development of IoT, the data generated by massive sensors and smart objects need to be processed in near real-time, which makes it extremely urgent to classify and detect abnormal data. 

In this paper, we explore the performance of the selected Deep Learning models for DAD tasks using the KDD Cup 99 dataset. Our research questions focus on the impact of different pre-precessing techniques, model architectures, optimisation algorithms, hyperparameters, ensemble techniques, and comparisons with other state-of-the-art methods on the performance of these models. Additionally, we aim to investigate the interpretability of the models and provide a comprehensive evaluation of their strengths and limitations.

\begin{table*}[h]
    \centering
    \resizebox{\linewidth}{!}{
    \begin{tabular}{p{6cm}p{3cm}p{3cm}p{4cm}}
    \hline
    \textbf{Paper Name and Reference} & \bfseries{Deep Algorithm Used} & \bfseries{Method Category} & \bfseries{Evaluation Metrics} \\ [1ex]
    \hline\hline
    Anomaly detection based on discriminative generative adversarial network\cite{appiah2021anomaly} & GAN & Reconstruction & Precision: 96.1\%, Recall: 97.2\%, F1: 96.9\% \\
    \hline
    Deep-compact-clustering based anomaly detection applied to electromechanical industrial systems\cite{arellano2021deep} & DAECC-OC-SVM & Clustering-based & Accuracy: 97.6\% \\
    \hline
    A deep learning approach for anomaly detection and prediction in power consumption data\cite{chahla2020deep} & LSTM & Predictability-based & Recall: 80\%, Accuracy: 89\%, F1:71\%, RMSE: 1.032 \\
    \hline
    A deep learning method for the detection and compensation of outlier events in stock data\cite{naidoo2022deep} & LSTM, HONN & Predictability-based & MAE: 0.03\% \\
    \hline
    A hybrid unsupervised clustering-based anomaly detection method\cite{pu2020hybrid} & OCSVM & Classification & Detection Rate: 89\% vs False Alarm Rate: 8\% :  \\
    \hline
    Unsupervised representation learning with deep convolutional generative adversarial networks\cite{radford2015unsupervised} & DCGAN & Reconstruction & Axccuracy: 82.8\% \\
    \hline
    A hybrid semi-supervised anomaly detection model for high-dimensional data\cite{song2017hybrid} & DAE & Reconstruction & AUC: 0.52 \\
    \hline
    Anomaly detection with convolutional neural networks for industrial surface inspection\cite{staar2019anomaly} & CNN & Distance-based & AUC: 0.83 \\
    \hline
    Distance-based anomaly detection for industrial surfaces using triplet networks\cite{tayeh2020distance} & CNN & Distance-based & AUC: an average class mean 0.093 higher \\
    \hline
    A deep learning enabled subspace spectral ensemble clustering approach for web anomaly detection\cite{yuan2017deep} & GMM, OCSVM (DEP-SSEC) & Clustering-based & DR: 92.3\%, FPR: 0.37\% \\
    \hline
    Deep Anomaly detection with self-supervised learning and adversarial training\cite{zhang2022deep} & DAT & Classification & F1: 87\%, AUROC: 97.15\%, AP: 75.40\%, ACC: 97.72\% \\
    \hline
    Network Anomaly Detection Based on Deep Support Vector Data Description\cite{chen2020network} & CNN, SVDD & Clustering-based & Accuracy: 96\%, Precision: 91.6\%, FPR: 6.7\%, FNR: 14\% \\
    \hline
    \end{tabular}
    }
    \caption{Research work on Deep Anomaly Detection algorithms}
    \label{tab: literature comparison}
\end{table*}
\section{Literature Review}
From a statistical perspective, most anomaly detection methods are based on constructing a probability distribution model and considering how likely the data can be fitted into the model\cite{wang2016anomaly}\cite{yang2015enhanced}. Therefore, anomalies are the data that have a low probability in the distribution model. To explore the suitable approaches to ensure network security and optimise performance, traditional and machine learning algorithms have been both adopted and developed for real-world applications. Moreover, Deep Learning (DL) methods for anomaly detection have been proven to show promising results in learning complex patterns and features from data and can detect anomalies in a relatively more accurate and efficient manner compared to traditional methods\cite{pecori2020iot}\cite{reddy2022network}. Traditional machine learning methods generally require more sophisticated feature engineering design, and the cross-domain versatility is not competitive \cite{schmidt2021deep}. Additionally, these methods are generally aimed at time series-based anomaly detection tasks, as the correlation between multiple series is difficult to design and compute by model-driven methods, while DL models are more suitable for such scenarios \cite{munir2019comparative}.

The relationship between deep learning and anomaly detection is inseparable because DL can be used for feature extraction or integrated with AD methods to learn effective representations of normal instances and even directly learn scalar anomaly scores in an end-to-end fashion\cite{pang2021deep}. AD algorithms based on deep learning methods can be mainly divided into the following categories: Distance-based, Classification, Clustering-based, Predictability-based, and Reconstruction method.

The DAD techniques have been popular due to the ability of deep neural networks to learn complex patterns and representations of data. As more DL algorithms have been proposed, a variety of choices for anomaly detection tasks are available nowadays, including autoencoder, deep belief networks, recurrent neural networks, convolutional neural networks, etc.

One advantage of adopting DL algorithms for anomaly detection is the ability to handle high-dimensional and complex data, which is beneficial in various real-world applications. This is because deep learning algorithms can learn complex data representations that capture local and global structures, and also handle both structured and unstructured data, making them well-suited for a wide range of scenarios \cite{chalapathy2019deep}. Additionally, DL algorithms can handle noisy data and identify anomalies accurately by learning robust representations of data. A comprehensive summary of recent research work on DAD algorithms can be found in Table 1. 

Despite the advantages we discussed, there are research gaps in the field of DAD that need to be addressed to further improve its effectiveness and efficiency. For example, DAD models often require a large amount of data for training, which can be a challenge for real-world applications with a limited amount of data. Also, DAD models are sensitive to outliers and typically designed to handle structured data, impacting the model's performance. Moreover, with a limited amount of labelled data, it is difficult to train DAD models, making adopting semi-supervised and unsupervised DAD models even more desirable.

\section{Research Approach \& Methodology}~\label{sec:design}

This work seeks to implement and assess existing algorithms from the literature within the context of deep anomaly detection. The primary objective is to execute these algorithms on benchmark datasets, with the ultimate goal of identifying potential enhancements. The methodology adopted in this research involves a meticulous and systematic exploration of deep anomaly detection.

The research will adopt a hypothesis-driven approach, postulating that deep learning (DL) models can proficiently identify anomalies within intricate datasets. The investigative process will encompass a combination of theoretical and empirical methods. These include an extensive literature review, thorough data pre-processing, and normalization procedures, as well as a rigorous implementation and training of models, followed by a robust evaluation and validation phase.

To ensure the credibility of the findings, the research outcomes will be subject to in-depth analysis using appropriate statistical methods. This statistical scrutiny aims to establish the statistical significance of the results. Moreover, ethical considerations will underpin the entire research process. This involves ensuring that data acquisition and utilization adhere strictly to ethical and legal standards. Overall, this work aspires to contribute valuable insights to the field of deep anomaly detection through a methodical and ethical research approach. 

We rely on deep learning models to detect anomalies, where the models are trained on the pre-processed dataset to identify patterns and anomalies in the data. The performance of the model is then evaluated using various metrics, such as accuracy precision, recall, and F1-score. The results are then analysed to determine the effectiveness of the DAD methods in solving the research problem. Finally, the findings are reported clearly and concisely, and the implications of the results are discussed in the context of the research question and objectives. We maintain a rigorous and systematic approach to ensure that the results are accurate and reliable.

\subsection{Popular Deep Learning Models for DAD}~\label{subsec:methodology}

\subsubsection{GAN}~\label{subsubsec: GAN}
 GANs are a class of deep learning algorithms that have two neural networks to be trained: a generator network and a discriminator network. The generator network learns to generate synthetic data that is similar to the normal data, while the discriminator network learns to distinguish between the synthetic data generated by the generator and the actual normal data. The generator and discriminator are trained together in a two-player minimax game, where the generator tries to fool the discriminator by generating synthetic data that is indistinguishable from the normal data, while the discriminator accurately identifies the synthetic data\cite{xia2022gan}.  

In the training process, the generator and discriminator networks are trained alternatively. First, the generator network is updated to generate synthetic data that is more similar to the normal data, and then the discriminator network is updated to better distinguish between the synthetic data and the normal data. The process continues until the generator network can generate synthetic data that is indistinguishable from the normal data, and the discriminator network cannot accurately distinguish between the two. After training, the discriminator network can be used to score new data points, with a lower score indicating a higher likelihood of being anomalous. Mathematically, the score for a new data point x can be defined as:
\begin{equation}
    S(x) = -log D(x)
\end{equation}

Strength. GANs are unsupervised learning algorithms, which means that they can detect anomalies without labels when the labelled data is scarce or difficult to obtain. They can also generate new data samples similar to the training data, which can be used in AD tasks to identify data that is significantly different from the normal data. Moreover, GANs are suitable for handling high-dimensional data, making them well-suited for complex datasets, such as KDD Cup 99.

Weaknesses. GANs can be difficult to train, as the training process can be unstable and prone to producing suboptimal results. Additionally, GANs are computationally intensive algorithms, which can be a drawback in real-world applications where the size of the dataset is large. These facts should be taken into account when processing our selected dataset.

\subsubsection{CNN}~\label{subsubsec: CNN}
CNNs are commonly used for image classification and anomaly detection in sequential data such as time series or network traffic data. CNNs are trained on the pre-processed and labelled data, learning to recognise patterns that are indicative of normal or anomalous behaviour. With the same advantage as RNNs that can identify complex patterns, CNNs can also learn and adapt to changes in the data over time, making them suitable for detecting anomalies in dynamic environments.

The convolution layer in a CNN model applies a convolution operation to the input data, which can be represented as 
\begin{equation}
    \textbf{C} = \textbf{X} * \textbf{W} + \textbf{b}
\end{equation}  
where $\textbf{C}$ is the output of the convolution layer, $\textbf{X}$ is the input data, $\textbf{W}$ is the convolution kernel, and $\textbf{b}$ is the bias. The convolution operation slides the convolution kernel over the input data and computes a dot product between the kernel and the overlapping region of the input data. This operation results in a feature map, which represents the most essential features in the data.

After the convolution operation, an activation function is applied to the output of the convolution layer. An activation function is used to introduce non-linearity into the network, allowing it to learn more complex data representations. A common activation function used in CNN is the rectified linear unit (ReLU) function, which is defined as: 
\begin{equation}
    f(x) = \max(0,x)
\end{equation}
After the activation function, a pooling layer is applied to reduce the size of the data, which helps to reduce overfitting and make the network computationally efficient. 

After multiple convolution and pooling layers, the data is fed into a fully connected layer. A fully connected layer is a type of layer in a neural network that connects all the neurons in one layer to all the neurons in the next layer. The fully connected layer can be mathematically represented as: 
\begin{equation}
    \textbf{Y} = \textbf{W}_2 \textbf{P} + \textbf{b}_2
\end{equation}
where $\textbf{Y}$ is the output of the fully connected layer, $\textbf{W}_2$ and $\textbf{b}_2$ are the weights and biases of the layer, and $\textbf{P}$ is the output of the pooling layer. In the last step, the data is fed into an output layer, which can be used to predict whether a given segment of IoT network traffic is anomalous or not \cite{ullah2021design}. 

Strength. CNN models can extract features from the data through convolutional and pooling layers, which can learn to recognise patterns in the dataset and reduce the dimensionality of the data. They also perform well and are robust to noise in detecting anomalies in sequential data, which is an efficient tool to characterise the differences between normal and anomalies.

Weakness. The complexity of CNN models makes them difficult to train and interpret and increases the computational requirements of the models. Another weakness of the CNN models is their tendency to overfit the training data, which results in poor performance on unseen data. Lastly, the CNN models require large amounts of high-quality data to train effectively, which can be a challenge when dealing with small datasets or corrupted data.

\subsubsection{LSTM}~\label{subsubsec: LSTM}
LSTM models are the type of recurrent neural network that is commonly used for DAD tasks. In an LSTM model, information is passed through a series of memory cells, gates, and layers to learn a temporal representation of the input data. This makes LSTM particularly well-suited for sequential data, where the order of the data points is important. A memory cell is the basic unit of an LSTM model. It holds information for a certain period of time, allowing the model to maintain its memory of past input even as new inputs are received. The memory cell is updated at each time step in the sequence \cite{laghrissi2021intrusion}.

Strengths. LSTM uses memory cells, which allow the network to store and access information from previous time steps. It makes LSTMs suitable for tasks that require the network to remember information from the past and use it to make predictions. LSTMs are also capable of handling long-term dependencies. Traditional RNNs struggle with this aspect because as the gap between relevant information and the current time step grows, the backpropagation gradient becomes weaker, and eventually vanishes. LSTMs solve this problem by using gates that control the flow of information into and out of the memory cells. In addition, LSTMs also solve the problems when the gradient used for backpropagation becomes too small to update the network weights effectively.

Weaknesses. One of the obvious weaknesses of LSTMs is their complexity, which requires a large number of parameters and computations, making it time-consuming and computationally expensive to train and deploy the infrastructure. LSTS are also prone to overfilling, making the model difficult to interpret.

\subsubsection{AutoEncoder}~\label{subsubsec: AE}
 AE is trained to reconstruct its input data. The idea behind using AEs for anomaly detection is to train the network on normal data, and then to identify deviations from the normal behaviours. 

The architecture of an autoencoder-based algorithm for DAD typically consists of two parts: an encoder and a decoder. The encoder maps the input data to a lower-dimensional representation, while the decoder maps the lower-dimensional representation back to the original data space. The training process minimises the reconstruction error between the input data and its reconstruction, which is obtained by passing the input data through the encoder and the decoder.

Once the AE is trained, it can be used for DAD by computing the reconstruction error for unseen data. If the reconstruction error for a particular instance is significantly larger than the reconstruction errors for the normal instances in the training set, it is considered to be an anomaly. The threshold for what constitutes a significant difference is typically set based on the distribution of the reconstruction errors for the normal instances in the training set \cite{zhou2017anomaly}.

Strengths. AEs are especially suitable for detecting anomalies in non-linear data, as they can learn complex, non-linear relationships between inputs and outputs. They can also be used for unsupervised AD tasks, and be adapted to different types of data, as they can be trained on a variety of data distributions, including high-dimensional and sparse data. Finally, AEs are generally robust to small amounts of noise in the data, as they are trained to reconstruct the input, rather than to classify it into a specific category.

Weaknesses. As with some other DL algorithms, AEs can also be computationally expensive, as they require a large number of training iterations and can be slow to converge. Also, if the dataset is small or the network architecture is complex, the AEs can be prone to overfitting. This also leads to the degradation of the model performance in the data, which is significantly different from the normal data, as the network may not be able to learn a good reconstruction for the anomalies.

\section{Experiments}~\label{sec:approach}
\subsection{Dataset}~\label{subsec:dataset}
We use KDD Cup’ 99 as our dataset. The dataset is a well-known benchmark dataset for intrusion detection, and it is most adopted for evaluating the performance of deep anomaly detection models. The dataset was created as part of the Knowledge Discovery and Data Mining (KDD) Cup competition held in 1999 and consists of a large set of network traffic data collected by the U.S. Defense Advanced Research Projects Agency (DARPA). It contains data from a simulated military network and includes various types of network attacks.

The dataset, consisting of over 4 million instances of network traffic, serves as an ideal choice for evaluating deep learning-based anomaly detection methods. Key considerations include its representatives of real-world IoT networks, ample size for training deep neural networks, pre-processing for data simplification, public availability for widespread research use, and continued relevance in addressing contemporary IoT security challenges due to its historical significance in network intrusion detection.

In a nutshell, the KDD Cup 1999 dataset is widely evaluated in the academic community and industry and is considered a benchmark dataset for evaluating the performance of anomaly detection algorithms. It is also a valuable resource for researchers studying network security and for practitioners building intrusion detection systems. A summary of the dataset can be found in Table 2. In our work, we will use the 10\% version of the KDD Cup 99 dataset to demonstrate experiment results.

\begin{table}[htbp]
    \centering
    \begin{tabular}{c c c c}
    \toprule
    \textbf{Data Type} & \textbf{Attack Name} & \textbf{Train set} & \textbf{Test set} \\ [1ex]
    \midrule
    Normal & -- & 77,815 & 19,463 \\
    Attack & DoS & 313,163 & 78,295 \\
    Attack & Prob & 3,284 & 823 \\
    Attack & R2L & 911 & 215 \\
    Attack & U2R & 43 & 9 \\ 
    \bottomrule
    \end{tabular}
    \caption{KDD Cup 99 dataset information}
    \label{tab:kdd99}
\end{table}

\subsection{Experiment Settings}~\label{subsec:implementaion}
\subsubsection{Data Pre-processing}~\label{subsubsec: pre-processing}
\begin{figure}[h]
    \centering
    \includegraphics[width=0.5\textwidth]{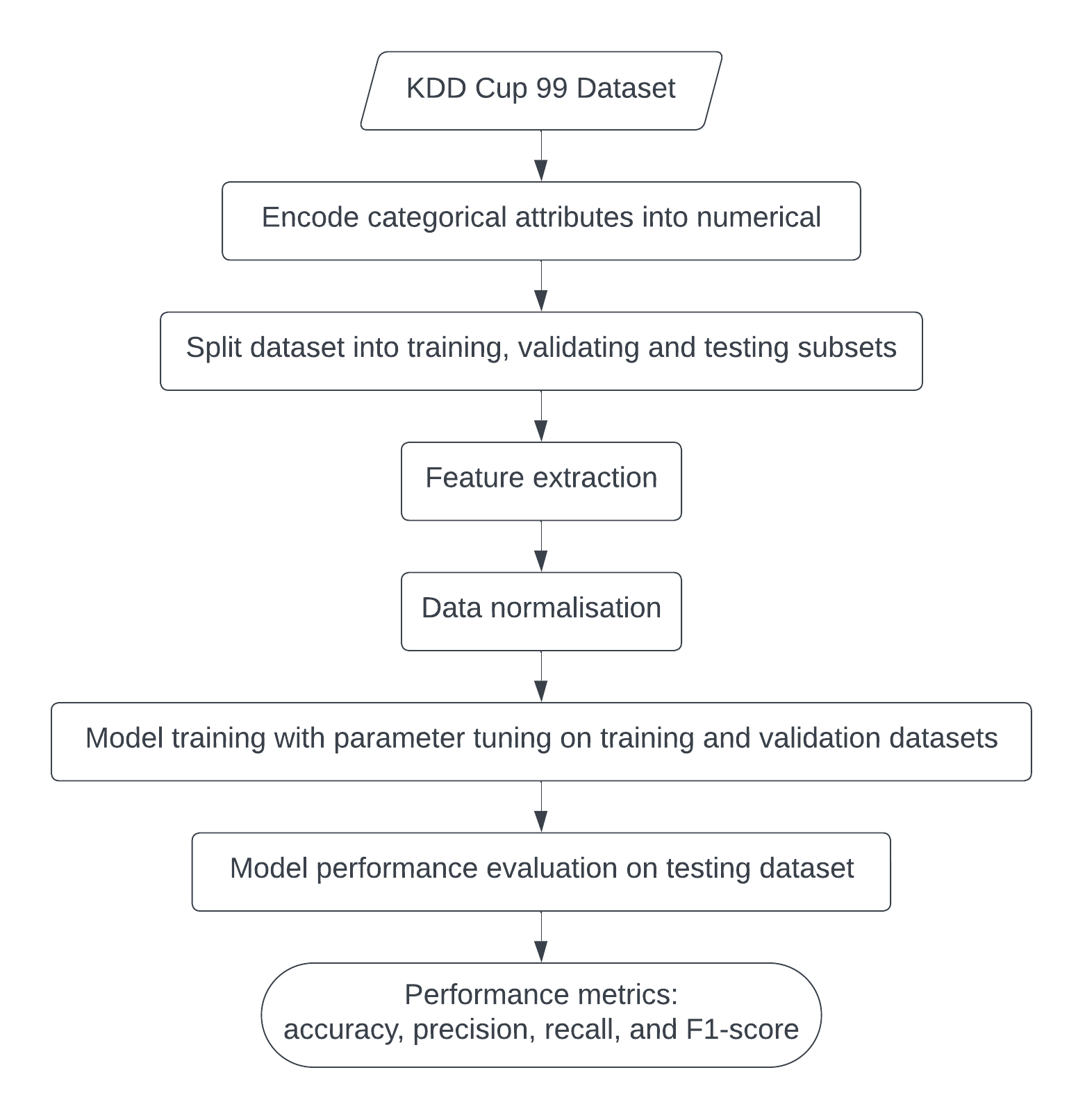}
    \caption{The illustration of the experimental process in our work}
    \label{fig:process}
\end{figure}

The original KDD Cup 99 dataset (10\%) contains 42 attributes and 494,021 entries. We first cleaned the dataset by dropping missing values and naming the attributes accordingly. We then encoded the categorical attributes (“protocol\_type”, “service”, and “flag”) and labels (“label”) and mapped the target variables into numerical values. We adopted both one-hot encoding and label encoding methods in our experiment, with One-hot encoding on the AE model, as it can be trained on sparse data, and labelling encoding is adopted for the rest models. 

Normaliser and MinMaxScaler from the sklearn package are used to transform the dataset after splitting it into training and testing. In the experiments, 80\% of the normal samples were randomly selected for training, and a testing dataset from the remaining samples was generated. We evaluate the performance of the models using metrics such as accuracy, precision, recall, and F1-score. Figure 2 gives an illustration of all the steps in our experimental design.

\begin{figure*}[h]
    \centering
    \includegraphics[width=0.9\textwidth]{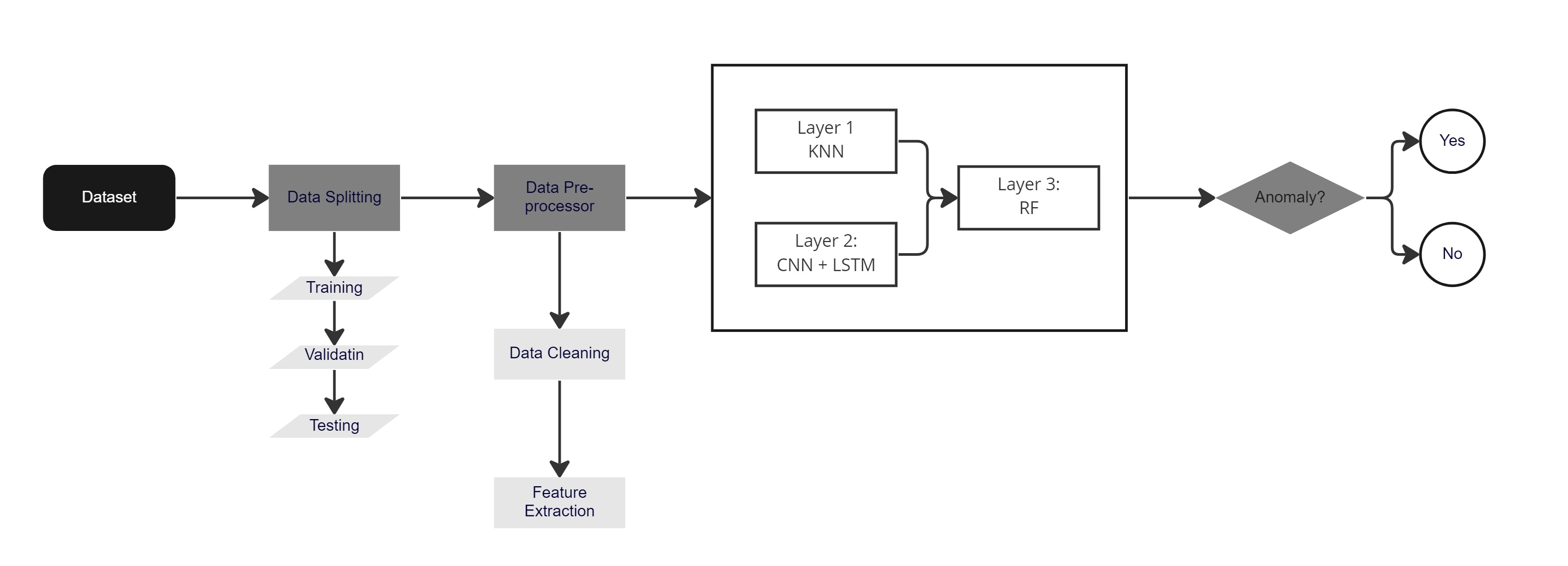}
    \caption{The flowchart of CNN+LSTM ensemble model.}
    \label{fig:ensemble flow}
\end{figure*}

\subsubsection{Model Implementation}~\label{subsubsec: settings}
We adopted a variety of pre-processing techniques, such as normalisation, feature scaling, and feature selection, to prepare the dataset. In the experiment phase, different hyperparameters have been varied for each model, such as the number of hidden layers, the size of the hidden layers, the learning rate, and the batch size. Then we evaluated the performance of each model using metrics to determine the optimal pre-processing techniques and hyperparameters that result in the best performance for each model. 

We implemented a GAN-based algorithm that incorporates an encoder that maps input samples to latent representations, as well as a generator and a discriminator during training. A score function is then defined to measure how unusual an example is based on a convex reconstruction loss and a combination of discriminator losses. The discriminator's cross-entropy or feature matching loss is used to evaluate whether the reconstructed data has similar features to the real samples in the discriminator. In the generator, we defined the latent space dimension of 114, with 6 hidden layers, a learning rate of $1e^{-5}$, and an activation function of “tanh”. For the discriminator, we defined a 6 hidden layer architecture with 128 neurons per layer, a “ReLU” activation function, a learning rate of 0.00001, and 0.2 dropout. For the GAN network, we trained for 10 epochs with a batch size of 512.

The other algorithm utilised the AE method as a discriminative DNN, where the target output is similar to the input, and the number of hidden layer nodes is lower than the input. Training is done by minimizing construction error and a regularization strategy, after which hidden layer results are considered as compressed representations of the input data. The method uses a deep autoencoder to encode features into different feature groups and uses the feature vector of the last hidden layer of the encoder as a representation of the attack score of the input data. The AE model is defined with one input layer, 3 hidden layers with activation functions of “tanh” and “ReLU”, and a decoder. 

In the proposed model, as illustrated in Figure 3, we adopted the CNN+LSTM method, which contains the network packet filters in three layers. The KNN is used for general categorisation, which classifies the input of the dataset. The second layer (CNN + LSTM) allows the model to analyse a series of data and makes the filter check the data against previously received packets. The outputs of the first two filters are then compared, and the conflicting input is sent to the third layer if a conflict is detected. The third layer is a Random Forest (RF) classifier. The RF classifier classifies the final result of the input. In this experiment, we adopted the “sparse\_categorical\_crossentropy” as the loss function and “SGD” as the optimizer. We also set the checkpoint by 1 verbose and monitored by validation accuracy, with overwriting the current file by the maximisation of the monitored quantity. When fitting the CNN + LSTM layers, we adopted 20 epochs and a batch size of 128. After training the RF classifier on conflicted instances, we combined the pre-trained models to make a final prediction on the unseen data.



\section{Results and Evaluations}~\label{sec:evaluation}

In our proposed model, the KNN layer produced an accuracy of 96.81\%, while the CNN+LSTM model obtained an accuracy of 97.83\% in overall detection. We then evaluated the assembled model by combining the three layers and gave the final output for the data. An accuracy of 98.22\% was acquired using the ensemble model, which is higher than any of the previous classifiers. We also compared the results among all three implemented models. The results can be found in Table 3.

 
\begin{table}[htbp]
    \centering
    \begin{tabular}{c c c c c}
    \toprule
    \textbf{Models} & \textbf{Accuracy} & \textbf{Precision} & \textbf{Recall} & \textbf{F1-score} \\ [1ex]
    \midrule
    Ensemble & 98.22\% & 92.67\% & 96.68\% & 96.21\% \\
    AE & 97.96\% & 90.68\% & 87.33\% & 93.45\% \\
    GAN & 90.28\% & 91.27\% & 92.86\% & 92.62\% \\ 
    \bottomrule
    \end{tabular}
    \caption{Implemented models results}
    \label{tab:result}
\end{table}

Taking all the performance metrics into account, the ensemble method best detected both normal and anomalous data instances. The confusion matrix advised that the number of false negatives and false positives was the lowest among all the models, indicating the model's ability to accurately identify both normal and anomalous data instances.

\section{Conclusion}
In this work, we conducted a literature review of deep anomaly detection on IoT network traffic analysis. The review has shown a growing interest in using DL methods for the detection of anomalies and highlighted several deep-learning models categorised by the nature of the methods. The result of this literature review provides a foundation for the following research on DAD analysis and highlights the potential of DL methods for the task. We then proposed to use ensemble techniques in current models for deep anomaly detection in IoT network traffic analysis. The model is evaluated in terms of accuracy, precision, recall, F1-score, and confusion matrix. 

The result of the empirical evaluation showed that the proposed model outperformed the existing models such as AE and GAN models in terms of precision, recall, and F1-score. The proposed model was able to achieve over 98\% accuracy, which is significantly higher than the other models. In light of this research's findings, we recommend using the ensemble model method for deep anomaly detection in IoT network traffic analysis and encourage future research in this area to improve further the performance of Deep Learning models for this task.






\bibliographystyle{IEEEtran}
%

\bibliography{reference}

\begin{thebibliography}{10}
\providecommand{\url}[1]{#1}
\csname url@samestyle\endcsname
\providecommand{\newblock}{\relax}
\providecommand{\bibinfo}[2]{#2}
\providecommand{\BIBentrySTDinterwordspacing}{\spaceskip=0pt\relax}
\providecommand{\BIBentryALTinterwordstretchfactor}{4}
\providecommand{\BIBentryALTinterwordspacing}{\spaceskip=\fontdimen2\font plus
\BIBentryALTinterwordstretchfactor\fontdimen3\font minus \fontdimen4\font\relax}
\providecommand{\BIBforeignlanguage}[2]{{%
\expandafter\ifx\csname l@#1\endcsname\relax
\typeout{** WARNING: IEEEtran.bst: No hyphenation pattern has been}%
\typeout{** loaded for the language `#1'. Using the pattern for}%
\typeout{** the default language instead.}%
\else
\language=\csname l@#1\endcsname
\fi
#2}}
\providecommand{\BIBdecl}{\relax}
\BIBdecl

\bibitem{chandola2009anomaly}
V.~Chandola, A.~Banerjee, and V.~Kumar, ``Anomaly detection: A survey. acm computing surveys,'' \emph{vol}, vol.~41, p.~15, 2009.

\bibitem{appiah2021anomaly}
B.~Appiah, Z.~Qin, O.~T. Nartey, B.~Agemang, and A.~J.~A. Kanpogninge, ``Anomaly detection based on discriminative generative adversarial network,'' \emph{International Journal of Network Security}, vol.~23, no.~4, pp. 718--724, 2021.

\bibitem{arellano2021deep}
F.~Arellano-Espitia, M.~Delgado-Prieto, A.-D. Gonzalez-Abreu, J.~J. Saucedo-Dorantes, and R.~A. Osornio-Rios, ``Deep-compact-clustering based anomaly detection applied to electromechanical industrial systems,'' \emph{Sensors}, vol.~21, no.~17, p. 5830, 2021.

\bibitem{chahla2020deep}
C.~Chahla, H.~Snoussi, L.~Merghem, and M.~Esseghir, ``A deep learning approach for anomaly detection and prediction in power consumption data,'' \emph{Energy Efficiency}, vol.~13, pp. 1633--1651, 2020.

\bibitem{naidoo2022deep}
V.~Naidoo and S.~Du, ``A deep learning method for the detection and compensation of outlier events in stock data,'' \emph{Electronics}, vol.~11, no.~21, p. 3465, 2022.

\bibitem{pu2020hybrid}
G.~Pu, L.~Wang, J.~Shen, and F.~Dong, ``A hybrid unsupervised clustering-based anomaly detection method,'' \emph{Tsinghua Science and Technology}, vol.~26, no.~2, pp. 146--153, 2020.

\bibitem{radford2015unsupervised}
A.~Radford, L.~Metz, and S.~Chintala, ``Unsupervised representation learning with deep convolutional generative adversarial networks,'' \emph{arXiv preprint arXiv:1511.06434}, 2015.

\bibitem{song2017hybrid}
H.~Song, Z.~Jiang, A.~Men, and B.~Yang, ``A hybrid semi-supervised anomaly detection model for high-dimensional data,'' \emph{Computational intelligence and neuroscience}, vol. 2017, 2017.

\bibitem{staar2019anomaly}
B.~Staar, M.~L{\"u}tjen, and M.~Freitag, ``Anomaly detection with convolutional neural networks for industrial surface inspection,'' \emph{Procedia CIRP}, vol.~79, pp. 484--489, 2019.

\bibitem{tayeh2020distance}
T.~Tayeh, S.~Aburakhia, R.~Myers, and A.~Shami, ``Distance-based anomaly detection for industrial surfaces using triplet networks,'' in \emph{2020 11th IEEE Annual Information Technology, Electronics and Mobile Communication Conference (IEMCON)}.\hskip 1em plus 0.5em minus 0.4em\relax IEEE, 2020, pp. 0372--0377.

\bibitem{yuan2017deep}
G.~Yuan, B.~Li, Y.~Yao, and S.~Zhang, ``A deep learning enabled subspace spectral ensemble clustering approach for web anomaly detection,'' in \emph{2017 International Joint Conference on Neural Networks (IJCNN)}.\hskip 1em plus 0.5em minus 0.4em\relax IEEE, 2017, pp. 3896--3903.

\bibitem{zhang2022deep}
X.~Zhang, J.~Mu, X.~Zhang, H.~Liu, L.~Zong, and Y.~Li, ``Deep anomaly detection with self-supervised learning and adversarial training,'' \emph{Pattern Recognition}, vol. 121, p. 108234, 2022.

\bibitem{chen2020network}
X.~Chen, C.~Cao, and J.~Mai, ``Network anomaly detection based on deep support vector data description,'' in \emph{2020 5th IEEE International Conference on Big Data Analytics (ICBDA)}.\hskip 1em plus 0.5em minus 0.4em\relax IEEE, 2020, pp. 251--255.

\bibitem{wang2016anomaly}
W.~Wang, B.~Zhang, D.~Wang, Y.~Jiang, S.~Qin, and L.~Xue, ``Anomaly detection based on probability density function with kullback--leibler divergence,'' \emph{Signal Processing}, vol. 126, pp. 12--17, 2016.

\bibitem{yang2015enhanced}
A.~Yang, J.~Zhang, L.~Pan, and Y.~Xiang, ``Enhanced twitter sentiment analysis by using feature selection and combination,'' in \emph{2015 International Symposium on Security and Privacy in Social Networks and Big Data (SocialSec)}.\hskip 1em plus 0.5em minus 0.4em\relax IEEE, 2015, pp. 52--57.

\bibitem{pecori2020iot}
R.~Pecori, A.~Tayebi, A.~Vannucci, and L.~Veltri, ``Iot attack detection with deep learning analysis,'' in \emph{2020 International Joint Conference on Neural Networks (IJCNN)}.\hskip 1em plus 0.5em minus 0.4em\relax IEEE, 2020, pp. 1--8.

\bibitem{reddy2022network}
D.~A. Reddy, V.~Puneet, S.~S.~R. Krishna, and S.~Kranthi, ``Network attack detection and classification using ann algorithm,'' in \emph{2022 6th International Conference on Computing Methodologies and Communication (ICCMC)}.\hskip 1em plus 0.5em minus 0.4em\relax IEEE, 2022, pp. 66--71.

\bibitem{schmidt2021deep}
B.~Schmidt and A.~Hildebrandt, ``Deep learning in next-generation sequencing,'' \emph{Drug discovery today}, vol.~26, no.~1, pp. 173--180, 2021.

\bibitem{munir2019comparative}
M.~Munir, M.~A. Chattha, A.~Dengel, and S.~Ahmed, ``A comparative analysis of traditional and deep learning-based anomaly detection methods for streaming data,'' in \emph{2019 18th IEEE international conference on machine learning and applications (ICMLA)}.\hskip 1em plus 0.5em minus 0.4em\relax IEEE, 2019, pp. 561--566.

\bibitem{pang2021deep}
G.~Pang, C.~Shen, L.~Cao, and A.~V.~D. Hengel, ``Deep learning for anomaly detection: A review,'' \emph{ACM computing surveys (CSUR)}, vol.~54, no.~2, pp. 1--38, 2021.

\bibitem{chalapathy2019deep}
R.~Chalapathy and S.~Chawla, ``Deep learning for anomaly detection: A survey,'' \emph{arXiv preprint arXiv:1901.03407}, 2019.

\bibitem{xia2022gan}
X.~Xia, X.~Pan, N.~Li, X.~He, L.~Ma, X.~Zhang, and N.~Ding, ``Gan-based anomaly detection: A review,'' \emph{Neurocomputing}, vol. 493, pp. 497--535, 2022.

\bibitem{ullah2021design}
I.~Ullah and Q.~H. Mahmoud, ``Design and development of a deep learning-based model for anomaly detection in iot networks,'' \emph{IEEE Access}, vol.~9, pp. 103\,906--103\,926, 2021.

\bibitem{laghrissi2021intrusion}
F.~Laghrissi, S.~Douzi, K.~Douzi, and B.~Hssina, ``Intrusion detection systems using long short-term memory (lstm),'' \emph{Journal of Big Data}, vol.~8, no.~1, p.~65, 2021.

\bibitem{zhou2017anomaly}
C.~Zhou and R.~C. Paffenroth, ``Anomaly detection with robust deep autoencoders,'' in \emph{Proceedings of the 23rd ACM SIGKDD international conference on knowledge discovery and data mining}, 2017, pp. 665--674.

\end{thebibliography}

\end{document}